%% file: main.tex
\documentclass[conference]{IEEEtran}
\IEEEoverridecommandlockouts
\usepackage{algorithm}
\usepackage{algorithmic}
\usepackage{amsmath,amssymb,amsfonts}
\usepackage{graphicx}
\usepackage{textcomp}
\usepackage{xcolor}
\usepackage{csquotes}
\usepackage{tikz}
\usepackage[nocompress]{cite}

\def\BibTeX{{\rm B\kern-.05em{\sc i\kern-.025em b}\kern-.08em
    T\kern-.1667em\lower.7ex\hbox{E}\kern-.125emX}}

\begin{document}

\title{Extend Wave Function Collapse Algorithm to Large-Scale Content Generation\\
{\footnotesize \textsuperscript{*} An algorithm framework that supports deterministic, aperiodic, infinite content generation that is suitable for level design}
% \thanks{This paper would like to thank ChatGPT for polishing the prose.}
}

\author{

\IEEEauthorblockN{1\textsuperscript{st} Yuhe NIE}
\IEEEauthorblockA{\textit{Computer Science and Engineering} \\
\textit{Southern University of Science and Technology}\\
Guangdong, China \\
nieyh2019@mail.sustech.edu.cn}

\and

\IEEEauthorblockN{2\textsuperscript{nd} Shaoming ZHENG}
\IEEEauthorblockA{\textit{Computer Science and Engineering} \\
\textit{Southern University of Science and Technology}\\
Guangdong, China \\
zhengsm2023@mail.sustech.edu.cn}

\and

\IEEEauthorblockN{3\textsuperscript{rd} Zhan ZHUANG}
\IEEEauthorblockA{\textit{Computer Science and Engineering} \\
\textit{Southern University of Science and Technology}\\
Guangdong, China \\
11811721@mail.sustech.edu.cn}

\and

\IEEEauthorblockN{4\textsuperscript{th} Xuan SONG$^{\dagger}$ \thanks{$\dagger$ Corresponding author}}
\IEEEauthorblockA{\textit{Computer Science and Engineering} \\
\textit{Southern University of Science and Technology}\\
Guangdong, China \\
songx@sustech.edu.cn}
}

\maketitle

\IEEEpubidadjcol

\input{Content/0_Abstract.tex}

\input{Content/1_Introduction.tex}

\input{Content/2_Related_Work.tex}

\input{Content/3_WFC_and_NWFC.tex}

\input{Content/4_Complete_and_SubComplete_Tilesets.tex}

\input{Content/5_Evaluation.tex}

\input{Content/6_Discussion.tex}

\input{Content/7_Conclusion.tex}

% How to use IEEE reference https://blog.csdn.net/kxbk100/article/details/88426028
\bibliographystyle{IEEEtran}
% \bibliography{ref}


\end{document}

%% file: Content/0_Abstract.tex
\begin{abstract}
Wave Function Collapse (WFC) is a widely used tile-based algorithm in procedural content generation, including textures, objects, and scenes. However, the current WFC algorithm and related research lack the ability to generate commercialized large-scale or infinite content due to constraint conflict and time complexity costs. This paper proposes a Nested WFC (N-WFC) algorithm framework to reduce time complexity. To avoid conflict and backtracking problems, we offer a complete and sub-complete tileset preparation strategy, which requires only a small number of tiles to generate aperiodic and deterministic infinite content. We also introduce the weight-brush system that combines N-WFC and sub-complete tileset, proving its suitability for game design. Our contribution addresses WFC's challenge in massive content generation and provides a theoretical basis for implementing concrete games.
\end{abstract}


%% file: Content/1_Introduction.tex
\section{Introduction}

\IEEEPARstart{V}{olume} of the game is growing. Developers are constantly looking for new ways to create immense virtual worlds. Procedural Content Generation (PCG) is a technique that can generate vast and varied game worlds without requiring developers to create every detail of the environment manually. Wave Function Collapse (WFC), proposed by Maxim Gumin\cite{gumin2016wfc}, is a popular PCG algorithm that can generate textures, 2D and 3D entities, and other abstract forms and is widely used in game\cite{freehold2014caveofqud, Stålberg2018badnorth, Stålberg2021townscraper}.

However, the current WFC algorithm suffers from exponential time complexity, which limits its practical application in large-scale or real-time infinite scene generation. As the size of the generated content increases, the time and number of conflicts WFC encounters will also increase, making the performance overhead worse. To address this challenge, this paper proposes a Nested Wave Function Collapse (N-WFC) algorithm framework that uses several internal WFCs (I-WFC) with fixed generation sizes nested within an exterior generation process. By maintaining constraints between adjacent content that I-WFC generates, N-WFC can generate large-scale or infinite scenes with polynomial time complexity.

To ensure that N-WFC generates accepted solutions, reduces conflicts, and avoids backtracking after I-WFC generation, we introduce the mathematical definitions of complete and sub-complete tilesets. The complete tileset requires a large number of tiles to satisfy the constraints exhaustively. In contrast, the sub-complete tileset can achieve similar effects using fewer tiles and is more practical for actual game development. We prove mathematically that both complete and sub-complete tilesets can generate infinite and aperiodic content. Importantly, every result generated by I-WFC under the N-WFC framework is deterministic and will not require backtracking.

We conduct a series of experiments using different sub-completions to test the efficacy of N-WFC. The results show that N-WFC significantly reduces the time required for content generation compared to the original WFC. Additionally, N-WFC is highly customizable and extensible, allowing for creating tilesets with different design properties using a weighted brush system. We demonstrate how the Carcassonne game can be used as an example to implement N-WFC and make design decisions.

In summary, this work solves the time complexity and conflict problems of WFC in content generation, enabling infinite, aperiodic, and deterministic content generation for both 2D and 3D game scene generation. N-WFC can be combined with other heuristics\cite{bailly2022genetic} or learning algorithms\cite{karth2021ml} to assist in the realization of design strategies, extending the performance and uses-scope of WFC and opening up new possibilities for procedural content generation.

%% file: Content/2_Related_Work.tex
\section{Related Work}

\subsection{PCG Algorithms for Scene Generation}

Procedural Content Generation (PCG) for scene generation has been a popular topic since the emergence of roguelike games \cite{roguelike_2023}. Rogue \cite{epyx1985rogue}, developed in 1985, was the first game that used PCG. Since then, numerous PCG techniques for scene generation have been proposed.

The goal of PCG for scenes is to add randomness to scene generation, making each generation different. However, the generated scene should still be logical and cater to the designer's ideas. As game developer Herbert Wolverson \cite{herbert2022makeroguelike} explained:

\begin{displayquote}
\textit{The randomness does not define the above games. It is fed into an algorithm that generates something that approximates what you want to get but ensures that it is different every time.}
\end{displayquote}

Early scene generation techniques focused on 2D dungeon-shaped games. Algorithms like Random Room Placement (RRP) or Binary Space Partition (BSP) \cite{baron2017dungeon} focused on the creation of logical and connected rooms. Later, Cellular Automata (CA) \cite{johnson2010autocell} and Drunkard Walk \cite{koesnaedi2022drunkard} were introduced to give the generated irregular scenes, making them look more like natural landscapes. Voronoi Diagrams \cite{aurenhammer2000voronoi} and Perlin Noise \cite{hart2001perlin} have pushed PCG towards generating larger city or continental plates. These algorithms use randomness and noise and add post-processing algorithms to ensure reasonable scenes. These techniques are suitable for 2D and 3D natural terrain generation.

However, these early algorithms were not suitable for strong-design-based scene generation. Maxim Gumin offered two algorithm strategies, the Wave Function Collapse (WFC) \cite{gumin2016wfc} algorithm and the MarkovJunior \cite{gumin2022markovjunior} algorithm, which made the generation result more of a product of design intuition than nature's randomness. In WFC, generation is described as a process of satisfying a binary constraint of interconnected nodes. In MarkovJunior, the technique is based on pattern matching and constraint propagation. Randomness no longer serves to generate shape but instead makes probabilistic choices based on design.

\subsection{Wave Function Collapse}

Wave Function Collapse (WFC) has gained significant attention since its introduction, as it generates output containing only patterns (tiles) similar to the input, making it suitable for generating objects for design purposes. Many game applications use WFC for small-range scene generation, such as Caves of Qud\cite{freehold2014caveofqud} with WFC-created ruins, Bad North\cite{Stålberg2018badnorth} for generating scenes and maintaining the navigation mesh for arranging the Viking war, and Townscraper\cite{Stålberg2021townscraper} for building a town underlying WFC mechanism. Additionally, some reinforcement learning teams\cite{team2021rlwfc} used WFC to generate arenas for their agents.

Several methodologies have been proposed to optimize the WFC algorithm, Paul C. Marrell \textit{et. al.} \cite{paul2021overlapping} expressed it as a Constraint Satisfy Problem (CSP) and applied the AC-3 algorithm to check the consistency. Karth \textit{et. al.}\cite{karth2022backtrackingandvq} examined backtracking and combined different heuristic algorithms in WFC, while Bailly \textit{et. al.}\cite{bailly2022genetic} mixed the genetic algorithm with WFC to guide generated content's novelty, complexity, and safety. Kim \textit{et. al.}\cite{kim2020graph} proposed a graph-based WFC that supports navigation in 3D scene generation. However, the problem of the exponential time complexity of the WFC algorithm still exists, limiting its ability to generate a large range of content. This problem also causes backtracking, making it unsuitable for commercial games. Marian Kleineberg\cite{marian2019infinitecity} tried to use WFC to create an experimental game, the \textit{Infinite City}, but he noted that conflicts and backtracking could lead to regenerating existing parts, further limiting its application in commercial games.

\subsection{PCG Serves as Level Design}

PCG is used in various aspects of game design, including level design, which primarily serves two purposes. First, it reduces the burden of creating different levels while maintaining the game's desired quality. Second, it creates varied experiences for players, encouraging them to use internal strategies to face unforeseen circumstances rather than relying on memorization.

PCG on level design abstracts the game mechanics, such as difficulty, rhythms, mood experiences, and task flows. There are several approaches to PCG on level design. \textit{Constructionist and Grammatist} algorithms use pre-defined content chunks, such as grammar rules, or randomly place chunks one after another. \textit{Constraint Driver}\cite{gumin2016wfc, gumin2022markovjunior} algorithms design specific constraints and use a constraint solver to find potential solutions. \textit{Optimizer and Learner}\cite{mourato2011geneticleveldesign, Khalifa2020PCGRL} algorithms respond to designing fitness functions and constraints of level features. They learn latent information and tune parameters independently to meet their desired outcomes.

For algorithms that assist in level design, whether explicit or implicit, they should expose controllable parameters and specific decision ideas for designers to use\cite{rabin2015pcgleveldesign}. These parameters can be used for various needs, such as curated design, player preferences, and adaptive difficulty.

%% file: Content/3_WFC_and_NWFC.tex
\section{WFC and N-WFC algorithm}

\subsection{WFC Introduction}
The Wave Function Collapse Algorithm (WFC) has two implementation strategies, the \textit{Simple Tiled Model} and the \textit{Overlapping Model}. These two models only differ in forming tilesets from input samples. Both of them have identical algorithm core, with the Overlapping Model generally regarded as more convenient due to its ability to extract tiles and their connectivity from the input sample image automatically.

\subsubsection{Simple Tiled Model}
\begin{figure}
\centering
\includegraphics[width=0.45\textwidth]{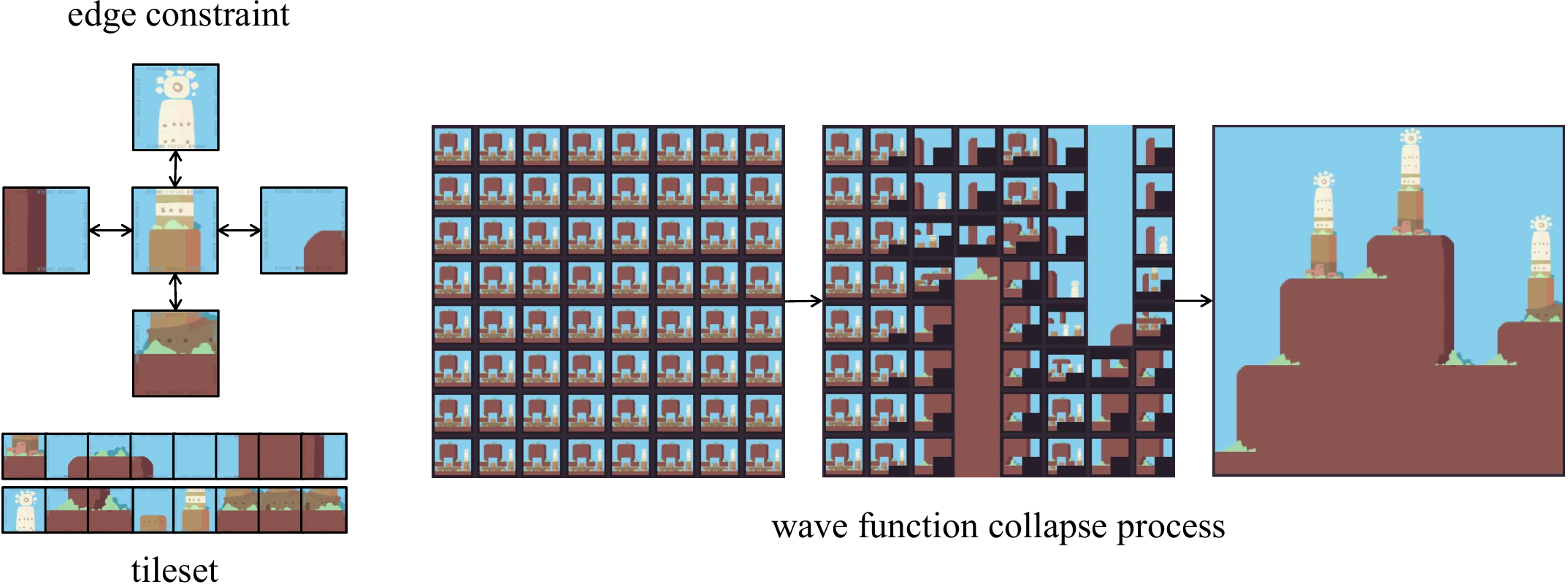}
\caption{Simple Tiled Model Demonstration (An interactive demo provided by Oskar Stålberg\cite{Stålberg2021wfcdemo}. The tileset contains the edge adjacencies information of each tile. WFC starts by creating a grid with unobservable (uncertain) cells. At each iteration, it picks one cell to collapse with one available tile for that cell that minimizes the Shannon entropy and propagates the edge constraints to affected cells. The algorithm terminates when all the cells have collapsed or a conflict is encountered)}
\label{fig:simple-tiled-model}
\end{figure}

The Simple Tiled Model can be either created manually or generated by the program. For a 2D game, as Fig. \ref{fig:simple-tiled-model} shows, it works by handcrafting the tileset and setting up the connectivity of the four edges of the tiles using binary constraints.

\subsubsection{Overlapping Model}
The Overlapping Model can automatically generate a tileset\footnote{Note that in Gumin's interpretation, the \textit{tiles} in an Overlapping Model are subdivided from the sample image, which is different from the Simple Tiled Model. While the unit of the ``tileset'' for the constraint solver is actually pattern region or group of tiles. Here we equate \textit{tiles} to pixels for simplicity.} from an input sample image. It extracts the tiles by using convolution to encode the pattern of the local neighborhood of each pixel on the sample image and then classifying them into tiles. The tiles adjacencies are also searched and constructed automatically.

\subsection{Problem Formulation}
This paper discusses WFC operates on the simple tiled model since the overlapping model doesn't meet the tileset requirements discussed later. For simplicity, the problem is formulated in the 2D case.

\subsubsection{Tileset and Edgeset}
Let $\mathcal{T}$ be a finite tileset. Each tile $t \in \mathcal{T}$ can be represented as a combination of $4$ edges. $t = (e_n, e_s , e_w, e_e)$ since edges are the only thing that matters in adjacency constraints. $e_n, e_s \in \mathcal{E_{NS}}$, $e_w, e_e \in \mathcal{E_{WE}}$. $\mathcal{E_{NS}}$ and $\mathcal{E_{WE}}$ are the sets of available edges on the north-south and west-east directions respectively. Depending on different implementations, $\mathcal{E_{WE}}$ and $\mathcal{E_{NS}}$ may or may not share some of the available edges. Each edge can be denoted as $e_{dir}(t),\ dir \in \{n,s,w,e\}$ for a tile $t$ or $e_{dir}(m, n), \ m \in [1,M], n \in [1,N]$ for a cell with index $(m,n)$.

\subsubsection{Initialization}
WFC executes on a $M\times N$ grid $\mathbf{G}$ where $g_{m,n} \subseteq \mathcal{T}$. A \textit{tiling} of $\mathbf{G}$ by $\mathcal{T}$ is a function $f: \mathbf{G} \leftarrow \mathcal{T}$. Initially, all the cells are in unobservable states. $\forall m \in [1, M], \forall n \in [1, N],\ g_{m,n} = \mathcal{T}$.

\subsubsection{WFC Solver}
The WFC solver starts looping by executing the following process:
\begin{enumerate}
    \item Choose a cell $g_{m,n}$ that has the minimum available tiles (Minimum Remaining Value).
    \item Randomly pick an available tile for the cell $t \in g_{m,n}$ and collapse this cell into a definite state: $g_{m,n} \leftarrow \{t\}$
    \item Propagate the new constraints into adjacency cells of $g_{i,j}$ and reduce tiles that do not satisfy the constraints until all the edge consistencies are maintained.
\end{enumerate}

\subsubsection{Accepted Result}
An \textit{accepted} WFC result is defined as\footnote{$e_{dir}^*$ denotes the equation will be checked if such cell exists.}:
\begin{flalign*}
    & \forall_{m=1}^M \forall_{n=1}^N |g_{m,n}| = 1\\
    & e_n(m, n) = e_s^*(m-1, n)\\
    & e_s(m, n) = e_n^*(m+1, n)\\
    & e_w(m, n) = e_e^*(m, n-1)\\
    & e_e(m, n) = e_w^*(m, n+1)
\end{flalign*}

\subsubsection{Time Complexity of WFC}
WFC is a constraint satisfaction problem that solves binary edge constraints, which may lead to conflicts during the generation process. It uses backtracking to ensure an accepted solution if it exists and uses AC-3 to optimize and maintain arc consistency. Suppose the number of the tileset is $|\mathcal{T}| = d$, the time complexity of WFC is $O(d^{M\times N} +  (M \times N)^2 d^3)$, which grows exponentially with the size of the generation grid.

\subsection{Nested Wave Function Collapse (N-WFC)}
The time-consuming problem of WFC can make it challenging to generate large-scale content. To address this issue, we propose the Nested Wave Function Collapse (N-WFC) algorithm, which splits a large-scale grid into smaller \textit{sub-grids} that can be solved more efficiently by using WFC to solve each sub-grid and maintaining the constraints between adjacent sub-grids. N-WFC is based on the domain splitting problem of CSP, which divides a large problem into smaller tasks while maintaining the overall constraints. Algorithm \ref{alg:n-wfc} shows the pseudocode of N-WFC, and the following sections explain how it works.

\subsubsection{Interior WFC (I-WFC)}
I-WFC generates a small and fixed-sized sub-grid $\mathbf{G}^{sub} \in \mathbb{Z}^2$ of size $C \times C$, where $C$ is a constant and typically a small value. I-WFC works similarly as regular WFC, with the only difference being that it may have some pre-constraints propagated from other sub-grids. Specifically, cells in the first row and first column of $\mathbf{G}^{sub}$ may have already been collapsed before starting the I-WFC: $\exists g^{sub}, \forall m \in [1, C], |g^{sub}_{m, 1}| = 1$, $\exists \mathbf{G}^{sub}, \forall n \in [1, C], |g^{sub}_{1, n}| = 1$.

\subsubsection{Exterior Generation Process}
To generate a N-WFC result, we split the large-scale grid into sub-grids problem and operated I-WFC to solve the problem. Suppose the entire task contains $A \times B$ sub-grids, the whole tasks contains $M  \times N$ cells, where $M = A\cdot(C-1) +1$, and $N = B\cdot (C-1) +1$. In the later section, we use $\mathbf{G}^{a, b}$ to denote for each sub-grid with index $(a, b)$. For simplicity, the generation process starts from the top-left corner and progresses in a diagonal sequence, layer by layer, which is called the \textit{diagonal generation process}. During generation, each sub-grid contains the value of its left neighbor's rightmost column and its upper neighbor's last row (if it exists), ensuring that the adjacent edges of adjacent sub-grids are consistent and can be treated as the same edge for overlapping. When all the sub-grids have been generated, each adjacent sub-grid overlaps its first row and first column, forming the N-WFC result.

\subsubsection{Accepted Result}
We denote each sub-grid as $\mathbf{G}^{a,b}, a \in [1, A], b \in [1, B]$. For the N-WFC framework, an \textit{accepted} N-WFC result will be\footnote{$(\mathbf{G}^{a,  b}_{m, n})^*$ denotes the equation will be checked if such sub-grid exists.}:
\begin{flalign*}
    & \forall_{a=1}^A \forall_{b=1}^B \mathbf{G}^{a,b} \text{ is an accepted WFC }\\
    & \forall_{a=1}^A \forall_{b=1}^B \forall_{m=1}^C g^{a,b}_{m, 1} = (g^{a,  b-1}_{m, C})^* \text{ and } |g^{a,b}_{m, 1}| = 1\\
    & \forall_{a=1}^A \forall_{b=1}^B \forall_{n=1}^Cg^{a,b}_{1, n} = (g^{a-1,b}_{C, n})^* \text{ and } |g^{a,b}_{m, 1}| = 1
\end{flalign*}

\begin{algorithm}  
    \caption{\textsc{Nested WFC}}
    \label{alg:n-wfc}  
    \begin{algorithmic}  
        \ENSURE returns an \textit{accepted} solution
        \STATE create a N-WFC grid $\mathbf{G}$
        \STATE $|\textbf{G}| = (A \cdot (C-1)+1) \times (B \cdot (C-1)+1)$
        \STATE split the $\textbf{G}$ into $A \times B$ sub-grid, each $|\mathbf{G}^{sub}| = C \times C$
        \FOR{layer in diagonal layer}
            \FOR{$\mathbf{G}^{a,b}$ in layer}
                \STATE initial empty sub-grid $\mathbb{G}, |\mathbb{G}| = C \times C$
                \STATE $\mathbb{G}_{:, 1} \leftarrow (\mathbf{G}^{a, b-1}_{:, C})^*$
                \STATE $\mathbb{G}_{1, :} \leftarrow (\mathbf{G}^{a-1, b}_{C, :})^*$
                \STATE $\mathbb{G} \leftarrow$ \textsc{WFC}($\mathbb{G}$)
                \STATE $\mathbf{G}^{a,b} = \mathbb{G}$
                \STATE paste $\mathbf{G}^{a,b}$ into $\textbf{G}$, overlapping one edge with $\mathbf{G}^{a-1,b}$ and $\mathbf{G}^{a,b-1}$ if exist
            \ENDFOR
        \ENDFOR
        \RETURN $\mathbf{G}$
    \end{algorithmic}  
\end{algorithm}

\subsubsection{Time Complexity of N-WFC}
N-WFC split the task into $A \times B$ small tasks. Each sub-grid uses I-WFC combining backtracking and AC-3. It has the time complexity of $O(d^{C^2} + C^4 d^3)$. The upper bound of N-WFC contains no more than $\frac{M \times N}{C^2}$ tasks. Therefore, the overall time complexity of N-WFC is $O(\frac{M \times N}{C^2} d^{C^2} + (M \times N) C^2 d^3)$. Since $C^2$ is a relatively small constant, N-WFC has only polynomial time complexity. As the range of generated content increases, The total computing time of N-WFC's will become significantly better than that of WFC.

\subsubsection{Conflict Problem}
While N-WFC has a better time complexity than WFC under ideal circumstances, it still faces the conflict problem caused by the edge constraints in practice. I-WFC may fail to find a suitable tile that satisfies certain pre-constraints propagated before, requiring the exterior generation process to use backtracking and has the probability of failing to generate an accepted solution. N-WFC is also less effective on the overlapping model, which constructs edge sets with a large amount of edge types, but with few tiles that satisfy the constraints. Therefore the overlapping model is not discussed in this paper. In Section \ref{sec:complete-and-sub-complete}, we introduce two tileset construction methods to solve the conflict problem.

%% file: Content/4_Complete_and_SubComplete_Tilesets.tex
\section{Complete and Sub-Complete Tilesets}
\label{sec:complete-and-sub-complete}

This section introduces two methods for creating tilesets that can be used for infinite, aperiodic tessellation as well as deterministic content generation. These methods require only a limited number of tiles and can be designed in 2D or 3D. The \textit{complete} tileset can accomplish tessellation without any conflicts during the WFC process, while the \textit{sub-complete} tileset can overcome conflicts using the N-WFC algorithm. We continue discussing them in the 2D form.

\subsection{Tileset Definition}
To maintain the aperiodic characteristic, we require $\max\{|\mathcal{E_{NS}}|, |\mathcal{E_{WE}}|\} \geq 2$, meaning that at least one of the edge sets contains two or more types of edges. 

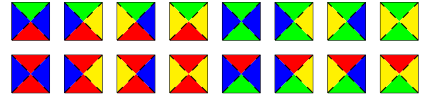
\begin{figure}
    \centering
    \begin{tikzpicture}
\draw (0.0, 0) -- (0.5, 0) -- (0.5, 0.5) -- (0.0, 0.5) -- (0.0, 0);
\fill[red] (0.0, 0.5)--(0.5, 0.5)--(0.25, 0.25)--(0.0, 0.5);
\fill[red] (0.0, 0)--(0.5, 0)--(0.25, 0.25)--(0.0, 0);
\fill[blue] (0.0, 0)--(0.0, 0.5)--(0.25, 0.25)--(0.0, 0);
\fill[blue] (0.5, 0.5)--(0.5, 0)--(0.25, 0.25)--(0.5, 0.5);
\draw[dashed] (0.0, 0)--(0.5, 0.5);
\draw[dashed] (0.5, 0)--(0.0, 0.5);
\draw (0.7, 0) -- (1.2, 0) -- (1.2, 0.5) -- (0.7, 0.5) -- (0.7, 0);
\fill[red] (0.7, 0.5)--(1.2, 0.5)--(0.95, 0.25)--(0.7, 0.5);
\fill[red] (0.7, 0)--(1.2, 0)--(0.95, 0.25)--(0.7, 0);
\fill[blue] (0.7, 0)--(0.7, 0.5)--(0.95, 0.25)--(0.7, 0);
\fill[yellow] (1.2, 0.5)--(1.2, 0)--(0.95, 0.25)--(1.2, 0.5);
\draw[dashed] (0.7, 0)--(1.2, 0.5);
\draw[dashed] (1.2, 0)--(0.7, 0.5);
\draw (1.4, 0) -- (1.9, 0) -- (1.9, 0.5) -- (1.4, 0.5) -- (1.4, 0);
\fill[red] (1.4, 0.5)--(1.9, 0.5)--(1.65, 0.25)--(1.4, 0.5);
\fill[red] (1.4, 0)--(1.9, 0)--(1.65, 0.25)--(1.4, 0);
\fill[yellow] (1.4, 0)--(1.4, 0.5)--(1.65, 0.25)--(1.4, 0);
\fill[blue] (1.9, 0.5)--(1.9, 0)--(1.65, 0.25)--(1.9, 0.5);
\draw[dashed] (1.4, 0)--(1.9, 0.5);
\draw[dashed] (1.9, 0)--(1.4, 0.5);
\draw (2.1, 0) -- (2.6, 0) -- (2.6, 0.5) -- (2.1, 0.5) -- (2.1, 0);
\fill[red] (2.1, 0.5)--(2.6, 0.5)--(2.35, 0.25)--(2.1, 0.5);
\fill[red] (2.1, 0)--(2.6, 0)--(2.35, 0.25)--(2.1, 0);
\fill[yellow] (2.1, 0)--(2.1, 0.5)--(2.35, 0.25)--(2.1, 0);
\fill[yellow] (2.6, 0.5)--(2.6, 0)--(2.35, 0.25)--(2.6, 0.5);
\draw[dashed] (2.1, 0)--(2.6, 0.5);
\draw[dashed] (2.6, 0)--(2.1, 0.5);
\draw (2.8, 0) -- (3.3, 0) -- (3.3, 0.5) -- (2.8, 0.5) -- (2.8, 0);
\fill[red] (2.8, 0.5)--(3.3, 0.5)--(3.05, 0.25)--(2.8, 0.5);
\fill[green] (2.8, 0)--(3.3, 0)--(3.05, 0.25)--(2.8, 0);
\fill[blue] (2.8, 0)--(2.8, 0.5)--(3.05, 0.25)--(2.8, 0);
\fill[blue] (3.3, 0.5)--(3.3, 0)--(3.05, 0.25)--(3.3, 0.5);
\draw[dashed] (2.8, 0)--(3.3, 0.5);
\draw[dashed] (3.3, 0)--(2.8, 0.5);
\draw (3.5, 0) -- (4.0, 0) -- (4.0, 0.5) -- (3.5, 0.5) -- (3.5, 0);
\fill[red] (3.5, 0.5)--(4.0, 0.5)--(3.75, 0.25)--(3.5, 0.5);
\fill[green] (3.5, 0)--(4.0, 0)--(3.75, 0.25)--(3.5, 0);
\fill[blue] (3.5, 0)--(3.5, 0.5)--(3.75, 0.25)--(3.5, 0);
\fill[yellow] (4.0, 0.5)--(4.0, 0)--(3.75, 0.25)--(4.0, 0.5);
\draw[dashed] (3.5, 0)--(4.0, 0.5);
\draw[dashed] (4.0, 0)--(3.5, 0.5);
\draw (4.2, 0) -- (4.7, 0) -- (4.7, 0.5) -- (4.2, 0.5) -- (4.2, 0);
\fill[red] (4.2, 0.5)--(4.7, 0.5)--(4.45, 0.25)--(4.2, 0.5);
\fill[green] (4.2, 0)--(4.7, 0)--(4.45, 0.25)--(4.2, 0);
\fill[yellow] (4.2, 0)--(4.2, 0.5)--(4.45, 0.25)--(4.2, 0);
\fill[blue] (4.7, 0.5)--(4.7, 0)--(4.45, 0.25)--(4.7, 0.5);
\draw[dashed] (4.2, 0)--(4.7, 0.5);
\draw[dashed] (4.7, 0)--(4.2, 0.5);
\draw (4.9, 0) -- (5.4, 0) -- (5.4, 0.5) -- (4.9, 0.5) -- (4.9, 0);
\fill[red] (4.9, 0.5)--(5.4, 0.5)--(5.15, 0.25)--(4.9, 0.5);
\fill[green] (4.9, 0)--(5.4, 0)--(5.15, 0.25)--(4.9, 0);
\fill[yellow] (4.9, 0)--(4.9, 0.5)--(5.15, 0.25)--(4.9, 0);
\fill[yellow] (5.4, 0.5)--(5.4, 0)--(5.15, 0.25)--(5.4, 0.5);
\draw[dashed] (4.9, 0)--(5.4, 0.5);
\draw[dashed] (5.4, 0)--(4.9, 0.5);

\draw (0.0, 0.7) -- (0.5, 0.7) -- (0.5, 1.2) -- (0.0, 1.2) -- (0.0, 0.7);
\fill[green] (0.0, 1.2)--(0.5, 1.2)--(0.25, 0.95)--(0.0, 1.2);
\fill[red] (0.0, 0.7)--(0.5, 0.7)--(0.25, 0.95)--(0.0, 0.7);
\fill[blue] (0.0, 0.7)--(0.0, 1.2)--(0.25, 0.95)--(0.0, 0.7);
\fill[blue] (0.5, 1.2)--(0.5, 0.7)--(0.25, 0.95)--(0.5, 1.2);
\draw[dashed] (0.0, 0.7)--(0.5, 1.2);
\draw[dashed] (0.5, 0.7)--(0.0, 1.2);
\draw (0.7, 0.7) -- (1.2, 0.7) -- (1.2, 1.2) -- (0.7, 1.2) -- (0.7, 0.7);
\fill[green] (0.7, 1.2)--(1.2, 1.2)--(0.95, 0.95)--(0.7, 1.2);
\fill[red] (0.7, 0.7)--(1.2, 0.7)--(0.95, 0.95)--(0.7, 0.7);
\fill[blue] (0.7, 0.7)--(0.7, 1.2)--(0.95, 0.95)--(0.7, 0.7);
\fill[yellow] (1.2, 1.2)--(1.2, 0.7)--(0.95, 0.95)--(1.2, 1.2);
\draw[dashed] (0.7, 0.7)--(1.2, 1.2);
\draw[dashed] (1.2, 0.7)--(0.7, 1.2);
\draw (1.4, 0.7) -- (1.9, 0.7) -- (1.9, 1.2) -- (1.4, 1.2) -- (1.4, 0.7);
\fill[green] (1.4, 1.2)--(1.9, 1.2)--(1.65, 0.95)--(1.4, 1.2);
\fill[red] (1.4, 0.7)--(1.9, 0.7)--(1.65, 0.95)--(1.4, 0.7);
\fill[yellow] (1.4, 0.7)--(1.4, 1.2)--(1.65, 0.95)--(1.4, 0.7);
\fill[blue] (1.9, 1.2)--(1.9, 0.7)--(1.65, 0.95)--(1.9, 1.2);
\draw[dashed] (1.4, 0.7)--(1.9, 1.2);
\draw[dashed] (1.9, 0.7)--(1.4, 1.2);
\draw (2.1, 0.7) -- (2.6, 0.7) -- (2.6, 1.2) -- (2.1, 1.2) -- (2.1, 0.7);
\fill[green] (2.1, 1.2)--(2.6, 1.2)--(2.35, 0.95)--(2.1, 1.2);
\fill[red] (2.1, 0.7)--(2.6, 0.7)--(2.35, 0.95)--(2.1, 0.7);
\fill[yellow] (2.1, 0.7)--(2.1, 1.2)--(2.35, 0.95)--(2.1, 0.7);
\fill[yellow] (2.6, 1.2)--(2.6, 0.7)--(2.35, 0.95)--(2.6, 1.2);
\draw[dashed] (2.1, 0.7)--(2.6, 1.2);
\draw[dashed] (2.6, 0.7)--(2.1, 1.2);
\draw (2.8, 0.7) -- (3.3, 0.7) -- (3.3, 1.2) -- (2.8, 1.2) -- (2.8, 0.7);
\fill[green] (2.8, 1.2)--(3.3, 1.2)--(3.05, 0.95)--(2.8, 1.2);
\fill[green] (2.8, 0.7)--(3.3, 0.7)--(3.05, 0.95)--(2.8, 0.7);
\fill[blue] (2.8, 0.7)--(2.8, 1.2)--(3.05, 0.95)--(2.8, 0.7);
\fill[blue] (3.3, 1.2)--(3.3, 0.7)--(3.05, 0.95)--(3.3, 1.2);
\draw[dashed] (2.8, 0.7)--(3.3, 1.2);
\draw[dashed] (3.3, 0.7)--(2.8, 1.2);
\draw (3.5, 0.7) -- (4.0, 0.7) -- (4.0, 1.2) -- (3.5, 1.2) -- (3.5, 0.7);
\fill[green] (3.5, 1.2)--(4.0, 1.2)--(3.75, 0.95)--(3.5, 1.2);
\fill[green] (3.5, 0.7)--(4.0, 0.7)--(3.75, 0.95)--(3.5, 0.7);
\fill[blue] (3.5, 0.7)--(3.5, 1.2)--(3.75, 0.95)--(3.5, 0.7);
\fill[yellow] (4.0, 1.2)--(4.0, 0.7)--(3.75, 0.95)--(4.0, 1.2);
\draw[dashed] (3.5, 0.7)--(4.0, 1.2);
\draw[dashed] (4.0, 0.7)--(3.5, 1.2);
\draw (4.2, 0.7) -- (4.7, 0.7) -- (4.7, 1.2) -- (4.2, 1.2) -- (4.2, 0.7);
\fill[green] (4.2, 1.2)--(4.7, 1.2)--(4.45, 0.95)--(4.2, 1.2);
\fill[green] (4.2, 0.7)--(4.7, 0.7)--(4.45, 0.95)--(4.2, 0.7);
\fill[yellow] (4.2, 0.7)--(4.2, 1.2)--(4.45, 0.95)--(4.2, 0.7);
\fill[blue] (4.7, 1.2)--(4.7, 0.7)--(4.45, 0.95)--(4.7, 1.2);
\draw[dashed] (4.2, 0.7)--(4.7, 1.2);
\draw[dashed] (4.7, 0.7)--(4.2, 1.2);
\draw (4.9, 0.7) -- (5.4, 0.7) -- (5.4, 1.2) -- (4.9, 1.2) -- (4.9, 0.7);
\fill[green] (4.9, 1.2)--(5.4, 1.2)--(5.15, 0.95)--(4.9, 1.2);
\fill[green] (4.9, 0.7)--(5.4, 0.7)--(5.15, 0.95)--(4.9, 0.7);
\fill[yellow] (4.9, 0.7)--(4.9, 1.2)--(5.15, 0.95)--(4.9, 0.7);
\fill[yellow] (5.4, 1.2)--(5.4, 0.7)--(5.15, 0.95)--(5.4, 1.2);
\draw[dashed] (4.9, 0.7)--(5.4, 1.2);
\draw[dashed] (5.4, 0.7)--(4.9, 1.2);

\end{tikzpicture}
    \caption{An Example of the Complete Tileset. It traverses all combinations of edge types and satisfies the minimum requirement of the complete tileset. $|\mathcal{T}| =  |\mathcal{E_{NS}}|^2 \cdot |\mathcal{E_{WE}}|^2 = 2^2 \cdot 2^2 = 16$.}
    \label{fig:complete}
\end{figure}

Fig. \ref{fig:complete} and Fig. \ref{fig:sub-complete} are examples of those tilesets. Specifically, we select four colors to represent different types of edges, $\mathcal{E_{NS}} = \text{[red, green]}$ and $\mathcal{E_{WE}} = \text{[blue, yellow]}$.

\subsubsection{Complete Tileset}
The complete tileset (Example as Fig. \ref{fig:complete}) is defined as:
$$\forall e_n, e_s \in \mathcal{E_{NS}}, \forall e_w, e_e \in \mathcal{E_{WE}}, \exists t \in \mathcal{T} \text{ s.t. } t = (e_n, e_s, e_w, e_e)$$
The complete tileset requires full permutation of all four edges. A complete tileset contains tiles large than: $|\mathcal{T}| \geq |\mathcal{E_{NS}}|^2 \cdot |\mathcal{E_{WE}}|^2$.

\subsubsection{Sub-Complete Tileset}
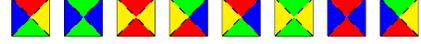
\begin{figure}
    \centering
    \begin{tikzpicture}
    \draw (0.0, 0.7) -- (0.5, 0.7) -- (0.5, 1.2) -- (0.0, 1.2) -- (0.0, 0.7);
\fill[red] (0.0, 1.2)--(0.5, 1.2)--(0.25, 0.95)--(0.0, 1.2);
\fill[green] (0.0, 0.7)--(0.5, 0.7)--(0.25, 0.95)--(0.0, 0.7);
\fill[blue] (0.0, 0.7)--(0.0, 1.2)--(0.25, 0.95)--(0.0, 0.7);
\fill[yellow] (0.5, 1.2)--(0.5, 0.7)--(0.25, 0.95)--(0.5, 1.2);
\draw[dashed] (0.0, 0.7)--(0.5, 1.2);
\draw[dashed] (0.5, 0.7)--(0.0, 1.2);
\draw (0.7, 0.7) -- (1.2, 0.7) -- (1.2, 1.2) -- (0.7, 1.2) -- (0.7, 0.7);
\fill[green] (0.7, 1.2)--(1.2, 1.2)--(0.95, 0.95)--(0.7, 1.2);
\fill[green] (0.7, 0.7)--(1.2, 0.7)--(0.95, 0.95)--(0.7, 0.7);
\fill[blue] (0.7, 0.7)--(0.7, 1.2)--(0.95, 0.95)--(0.7, 0.7);
\fill[blue] (1.2, 1.2)--(1.2, 0.7)--(0.95, 0.95)--(1.2, 1.2);
\draw[dashed] (0.7, 0.7)--(1.2, 1.2);
\draw[dashed] (1.2, 0.7)--(0.7, 1.2);
\draw (1.4, 0.7) -- (1.9, 0.7) -- (1.9, 1.2) -- (1.4, 1.2) -- (1.4, 0.7);
\fill[red] (1.4, 1.2)--(1.9, 1.2)--(1.65, 0.95)--(1.4, 1.2);
\fill[red] (1.4, 0.7)--(1.9, 0.7)--(1.65, 0.95)--(1.4, 0.7);
\fill[yellow] (1.4, 0.7)--(1.4, 1.2)--(1.65, 0.95)--(1.4, 0.7);
\fill[yellow] (1.9, 1.2)--(1.9, 0.7)--(1.65, 0.95)--(1.9, 1.2);
\draw[dashed] (1.4, 0.7)--(1.9, 1.2);
\draw[dashed] (1.9, 0.7)--(1.4, 1.2);
\draw (2.1, 0.7) -- (2.6, 0.7) -- (2.6, 1.2) -- (2.1, 1.2) -- (2.1, 0.7);
\fill[green] (2.1, 1.2)--(2.6, 1.2)--(2.35, 0.95)--(2.1, 1.2);
\fill[red] (2.1, 0.7)--(2.6, 0.7)--(2.35, 0.95)--(2.1, 0.7);
\fill[yellow] (2.1, 0.7)--(2.1, 1.2)--(2.35, 0.95)--(2.1, 0.7);
\fill[blue] (2.6, 1.2)--(2.6, 0.7)--(2.35, 0.95)--(2.6, 1.2);
\draw[dashed] (2.1, 0.7)--(2.6, 1.2);
\draw[dashed] (2.6, 0.7)--(2.1, 1.2);
\draw (2.8, 0.7) -- (3.3, 0.7) -- (3.3, 1.2) -- (2.8, 1.2) -- (2.8, 0.7);
\fill[red] (2.8, 1.2)--(3.3, 1.2)--(3.05, 0.95)--(2.8, 1.2);
\fill[green] (2.8, 0.7)--(3.3, 0.7)--(3.05, 0.95)--(2.8, 0.7);
\fill[yellow] (2.8, 0.7)--(2.8, 1.2)--(3.05, 0.95)--(2.8, 0.7);
\fill[blue] (3.3, 1.2)--(3.3, 0.7)--(3.05, 0.95)--(3.3, 1.2);
\draw[dashed] (2.8, 0.7)--(3.3, 1.2);
\draw[dashed] (3.3, 0.7)--(2.8, 1.2);
\draw (3.5, 0.7) -- (4.0, 0.7) -- (4.0, 1.2) -- (3.5, 1.2) -- (3.5, 0.7);
\fill[green] (3.5, 1.2)--(4.0, 1.2)--(3.75, 0.95)--(3.5, 1.2);
\fill[green] (3.5, 0.7)--(4.0, 0.7)--(3.75, 0.95)--(3.5, 0.7);
\fill[yellow] (3.5, 0.7)--(3.5, 1.2)--(3.75, 0.95)--(3.5, 0.7);
\fill[yellow] (4.0, 1.2)--(4.0, 0.7)--(3.75, 0.95)--(4.0, 1.2);
\draw[dashed] (3.5, 0.7)--(4.0, 1.2);
\draw[dashed] (4.0, 0.7)--(3.5, 1.2);
\draw (4.2, 0.7) -- (4.7, 0.7) -- (4.7, 1.2) -- (4.2, 1.2) -- (4.2, 0.7);
\fill[red] (4.2, 1.2)--(4.7, 1.2)--(4.45, 0.95)--(4.2, 1.2);
\fill[red] (4.2, 0.7)--(4.7, 0.7)--(4.45, 0.95)--(4.2, 0.7);
\fill[blue] (4.2, 0.7)--(4.2, 1.2)--(4.45, 0.95)--(4.2, 0.7);
\fill[blue] (4.7, 1.2)--(4.7, 0.7)--(4.45, 0.95)--(4.7, 1.2);
\draw[dashed] (4.2, 0.7)--(4.7, 1.2);
\draw[dashed] (4.7, 0.7)--(4.2, 1.2);
\draw (4.9, 0.7) -- (5.4, 0.7) -- (5.4, 1.2) -- (4.9, 1.2) -- (4.9, 0.7);
\fill[green] (4.9, 1.2)--(5.4, 1.2)--(5.15, 0.95)--(4.9, 1.2);
\fill[red] (4.9, 0.7)--(5.4, 0.7)--(5.15, 0.95)--(4.9, 0.7);
\fill[blue] (4.9, 0.7)--(4.9, 1.2)--(5.15, 0.95)--(4.9, 0.7);
\fill[yellow] (5.4, 1.2)--(5.4, 0.7)--(5.15, 0.95)--(5.4, 1.2);
\draw[dashed] (4.9, 0.7)--(5.4, 1.2);
\draw[dashed] (5.4, 0.7)--(4.9, 1.2);
\end{tikzpicture}
    \caption{An example of the Sub-complete Tileset. Color of the edge represents different edge types. It requires full permutation of $(e_n, e_s)$, $(e_w, e_e)$, $(e_n, e_w)$, and $(e_s, e_e)$ respectively. In fact, the first four tiles have formed a minimum sub-complete tileset in this case. Here we add more tiles that also satisfy the requirements.}
    \label{fig:sub-complete}
\end{figure}

The sub-complete tileset (Example as Fig. \ref{fig:sub-complete}) is defined as:
\begin{flalign*}
    & \forall e_n, e_s \in \mathcal{E_{NS}}, \exists t \in \mathcal{T} \text{ s.t. } e_n(t) = e_n \wedge e_s(t) = e_s\\
    & \forall e_w, e_e \in \mathcal{E_{WE}},  \exists t \in \mathcal{T} \text{ s.t. } e_w(t) = e_w \wedge e_e(t) = e_e\\
    & \forall e_n \in \mathcal{E_{NS}}, \forall e_w \in \mathcal{E_{WE}},  \exists t \in \mathcal{T} \text{ s.t. } e_n(t) = e_n \wedge e_w(t) = e_w\\
    & \forall e_s \in \mathcal{E_{NS}}, \forall e_e \in \mathcal{E_{WE}},  \exists t \in \mathcal{T} \text{ s.t. } e_s(t) = e_s \wedge e_e(t) = e_e
\end{flalign*}

The sub-complete tileset requires creating tiles for at least: $|\mathcal{T}| \geq \max\{|\mathcal{E_{NS}}|^2, |\mathcal{E_{WE}}|^2\}$. There are many combinations of sub-complete tilesets. If two tilesets  $\mathcal{T}_1$, $\mathcal{T}_2$ have the same edge sets $\mathcal{E_{NS}}$ and $\mathcal{E_{WE}}$, $\mathcal{T}_1$ is a sub-complete tileset,  $\mathcal{T}_2$ is a complete tileset, then $\mathcal{T}_1 \subseteq \mathcal{T}_2$.

The sub-complete tileset meets all the characteristics of the complete tileset. However, under the edge set of the same size, the number of tiles required by the sub-complete tileset is much smaller than that of the complete tileset, making it a more practical option for game production.

\subsection{Infinity}
\label{sec:infinity-characteristic}
The infinity characteristic ensures algorithm will always return an accepted solution regardless of the range of the problem.

\subsubsection{Definition}
An \textit{infinity} tiling  $f: \mathbf{G} \leftarrow \mathcal{T}$ is:
\begin{flalign*}
    & \forall_{m=1}^\infty \forall_{n=1}^\infty |g_{m,n}| = 1\\
    & e_n(m, n) = e_s^*(m-1, n) \wedge e_s(m, n) = e_n^*(m+1, n)\\
    & e_w(m, n) = e_s^*(m, n-1) \wedge e_e(m, n) = e_w^*(m, n+1)
\end{flalign*}

\subsubsection{Infinity of Complete Tileset}
It is easy to prove that the complete tileset can satisfy the definition. For a cell $g_{m, n}$, the strictest constraint a tile $t \in \mathcal{T}$ should satisfy is that the edge constraints propagate its four adjacent collapsed sub-grids, such that:
\begin{flalign*}
    & e_n(t) = e_s(m-1, n) \wedge e_s(t) = e_n(m+1, n)\\
    & e_w(t) = e_e(m, n-1) \wedge e_e(t) = e_w(m, n+1)
\end{flalign*}

According to the definition of the complete tileset. We can always find at least one tile to satisfy such constraints.

\subsubsection{Infinity of Sub-complete Tileset}
To prove the infinity on the sub-complete tileset. We can use the diagonal generation process as mentioned in N-WFC. Suppose the infinity grid $\mathbf{G}$ starts from the top-left corner $\mathbf{G}^{1,1}$, and the generation sequence of each tile follows the diagonal generation process. In this process, the most edge constraints a tile will encounter are when it is set with two edges already constrained by its upper and left adjacent tiles. For a cell $g_{m, n}$, we need to find a tile $t \in \mathcal{T}$, such that:
$$e_n(t) = e_s(m-1, n) \wedge e_w(t) = e_e(m, n-1)$$
According to one definition of the sub-complete tileset:
$$\forall e_n \in \mathcal{E_{NS}}, \forall e_w \in \mathcal{E_{WE}},  \exists t \in \mathcal{T} \text{ s.t. } e_n(t) = e_n \wedge e_w(t) = e_w$$
We can always find at least one tile that satisfies such constraints.

\subsection{Aperiodicity}
The aperiodicity characteristic ensures the algorithm generates non-repetitive content, meaning each generation has a different result, making the game more Rogue-like.

\subsubsection{Definition}
A tiling is called \textit{periodic} with period $(a, b) \in \mathbb{Z}^2$  $f: \mathbf{G} \leftarrow \mathcal{T}$ if
$$\forall (m, n) \in \mathbb{Z}^2: |g_{m,n}| = 1 \wedge g_{m,n} = g_{m+a,n+b}$$
$f$ is called \textit{aperiodicity} if $f$ is not periodic.

\subsubsection{Aperiodicity of Complete and Sub-Complete Tileset}
Suppose the generation process on $\mathbf{G}$ still uses diagonal generation. Consider an arbitrary cell $g_{m, n}$ that has not been assigned a tile yet. By the edge constraints, this cell has two neighbors $g_{m, n-1}, g_{m-1, n}$ to the north and west that have already been assigned tiles. Note that if $m=1$ or $n=1$, the corresponding neighbor may not exist, but that is not a problem as we can assume that such neighbors are assigned an arbitrary tile that satisfies the missing constraint.

Either $\mathcal{T}$ is complete or sub-complete, according to the definition, there exist at least two tiles $t \in \mathcal{T}$ that satisfy the two edge constraints $e_n(t) = e_s(m-1, n), e_w(t) = e_e(m, n-1)$. Thus, at least two tiles choices are to be selected by $g_{m,n}$. It does not repeat previous patterns, which causes an independent random process. This argument applies to any uncollapsed cell, making the tiling process $f$ using the complete or sub-complete tileset an aperiodic tiling.

\subsection{Determinacy}
The characteristic of \textit{determinacy} is a context definition. It is defined based on combining the use of N-WFC and sub-complete tileset. While the fine-grained diagonal generation process rather than I-WFC may be used directly for the complete and sub-complete tileset, the combination of N-WFC and sub-complete tilesets is more suitable for design purposes. We will discuss this in section \ref{sec:evaluation}.

The N-WFC framework is well-suited for infinite content generation. In practice, \textit{infinity} means generating scenes before the user reaches them. Marian Kleinebery\cite{marian2019infinitecity}, the creator of \textit{infinite city}, mentioned that WFC has the problem that some places, although shown to the player, require backtracking and re-generation due to conflicts.
\begin{displayquote}
\textit{In my opinion, this limitation makes the WFC approach for infinite worlds unsuitable for commercial games.}
\end{displayquote}
Determinacy is the characteristic that directly solves this problem. We prove that each sub-grid N-WFC has generated will never be backtracked during the whole generation process.

\subsubsection{Definition}
A N-WFC tiling $f: \mathbf{G} \leftarrow \mathcal{T}, \mathcal{T}$ is called \textit{determinacy} if
\begin{flalign*}
    & \forall_{a=1}^A \forall_{b=1}^B f: \mathbf{G}^{a,b} \leftarrow \mathcal{T} \text{ is an accepted I-WFC result}\\
    & \mathbf{G}^{a,b} \text{ will never be backtracked}
\end{flalign*}

\subsubsection{Determinacy of Sub-complete Tileset}
For N-WFC, The strongest constraint is $\exists \mathbf{G}^{sub}, \forall m \in [1, C] , |g^{sub}_{m, 1}| = 1$, $\exists \mathbf{G}^{sub}, \forall n \in [1, C], |g^{sub}_{1, n}| = 1$. As we break $\mathbf{G}^{sub}$ into cells, all the pre-constraints are related to $e_n(t)$ and $e_w(t)$. As proved in section \ref{sec:infinity-characteristic}, the diagonal generation process always produces one solution for a sub-grid. Since I-WFC uses the backtracking strategy, it is guaranteed to return an accepted solution if one exists. Therefore, N-WFC with the sub-complete tileset is deterministic.

%% file: Content/5_Evaluation.tex
\section{Experiment and Evaluation}
\label{sec:evaluation}
\subsection{Time Computing between N-WFC with Sub-complete Tileset and Traditional WFC}
We evaluated the sub-complete tileset's performance using the N-WFC framework and compared it with the traditional WFC algorithm. Both of them used the Minimum Remaining Value strategy for wave observation and Random Selection Strategy for wave collapse. In the exterior of the N-WFC, we used the diagonal generation process.

To evaluate the time complexity, we created a series of sub-complete tilesets with different numbers of edge sets. We assigned the value $C$ a small constant for I-WFC and evaluated the total computation time in different grid sizes. For each grid size that used different tilesets, the experiment was executed $100$ times, and we calculated the mean and variance of the overall time in finding an accepted solution.

\begin{figure*}
\centering
\includegraphics[width=0.95\textwidth]{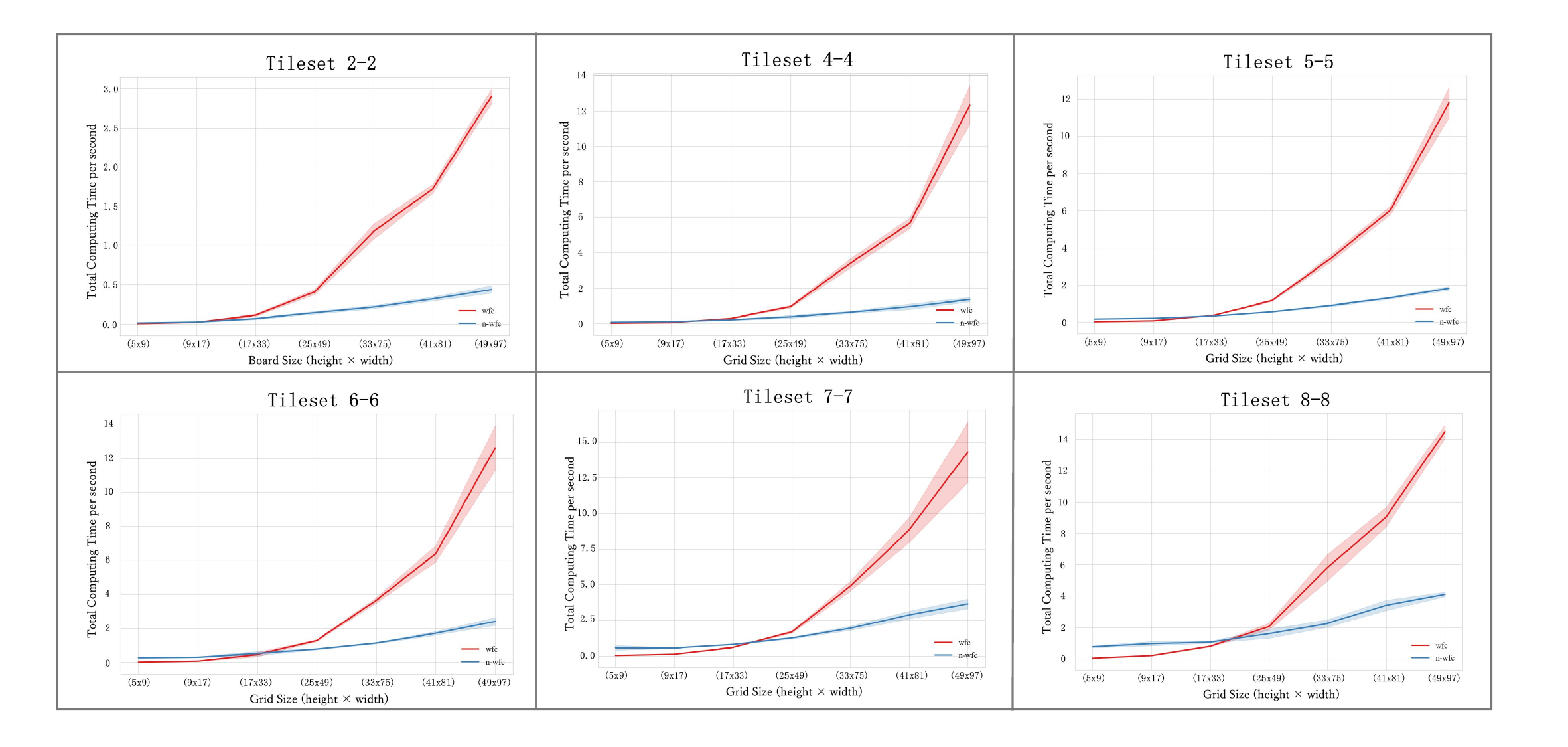}
\caption{Total computation time between WFC and N-WFC in 6 different sub-complete tilesets. The horizontal axis is Grid Size, and the vertical axis is the average computation time (in seconds) required to generate an accepted solution.}
\label{fig:NWFC&WFC}
\end{figure*}

Fig. \ref{fig:NWFC&WFC}. shows the result of the experiment. We created $6$ different sub-complete tilesets with different sizes of edge sets. For simplicity, we set the edge set $|\mathcal{E_{NS}}| = |\mathcal{E_{WE}}| = \{2,3,4,5,6,7\}$, and we assigned $C = 5$. We evaluated the total computation time per second through different grid sizes $[(5\times 9), (9 \times 17), (17 \times 33), (25 \times 49), (33 \times 75), (41 \times 81), (49 \times 97)]$. We didn't test other larger grids because WFC took too long to finish.

The results show that the time that WFC spent grows exponentially, while N-WFC grows in polynomials. For a large-scale content generation, N-WFC only uses I-WFC to backtrack the fixed range of sub-grids, outperforming traditional WFC.

\subsection{Implementation Example: Carcassonne}
As an example of deterministic large-scale or infinite content generation, we used the board game Carcassonne, which provides logically four edge types: $e_0$ for grass, $e_1$ for city, $e_2$ for path, and $e_3$ for stream. Specifically, in this game, $\mathcal{E_{NS}} = \mathcal{E_{WE}} = \{e_0, e_1, e_2, e_3\}$.  We created a sub-complete tileset that satisfies the requirements as shown in Fig. \ref{fig:Carcassonne-example}-a. It consists of several tiles as shown in Table \ref{tab:carcassonne-sub-complete}. The sub-complete tileset we're using is a very small subset. In the actual implementation of the game, tiles can be either symmetry or rotated as needed. Multiple tiles with identical logical forms are also accepted, providing more art asset combinations.

\subsubsection{Large-Scale Content}
\begin{figure*}
\centering
\includegraphics[width=0.95\textwidth]{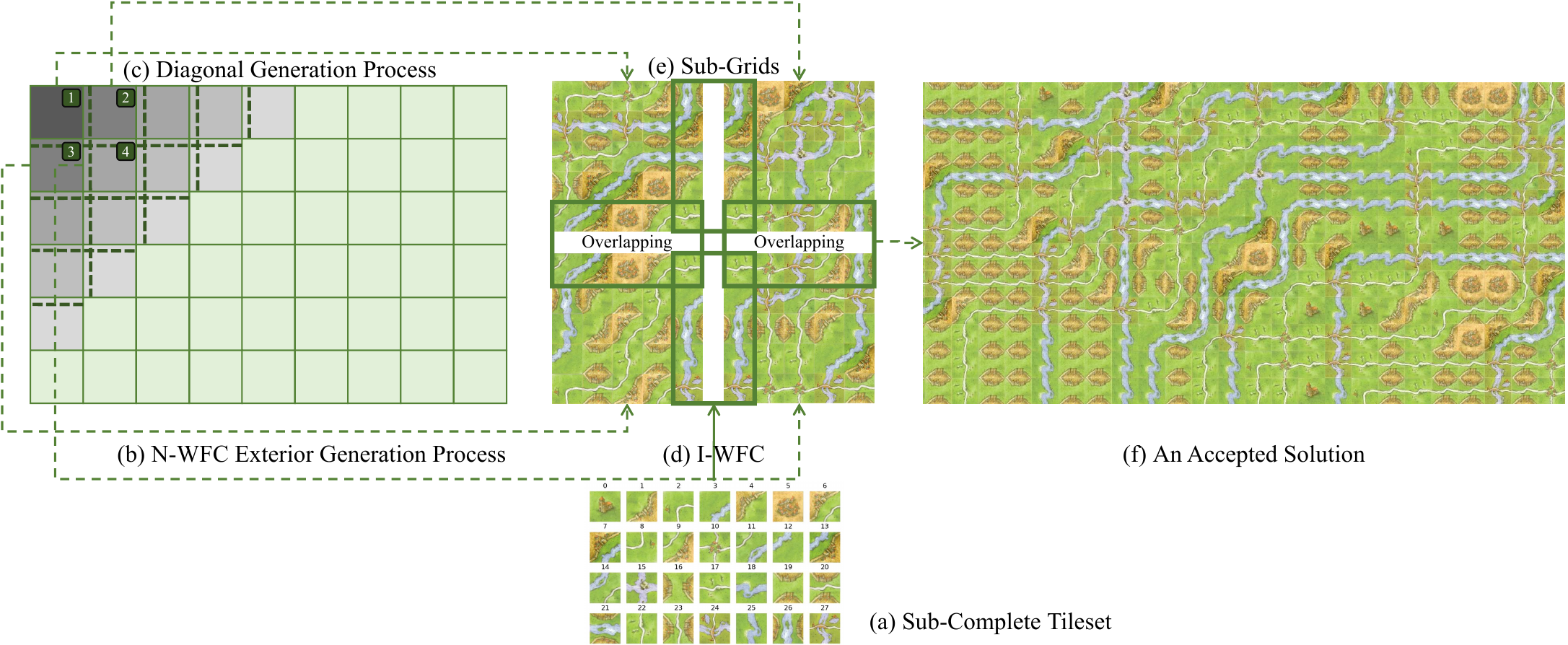}
\caption{Large-scale game implementation with N-WFC and sub-complete tileset. First, it requires (a) one sub-complete tileset. Then the (b) Exterior Generation Process uses (c) Diagonal Generation Process to start generating. Each (d) sub-grid uses (e) I-WFC to find an accepted solution and overlap its edge with the adjacent sub-grids, forming an (f) final soluton.}
\label{fig:Carcassonne-example}
\end{figure*}

To generate a large-scale scene, we set the size of the grid to be generated by N-WFC (Fig. \ref{fig:Carcassonne-example}-b) and passed the constraints of the adjacent sub-grids to the I-WFC generation process in a diagonal order (Fig. \ref{fig:Carcassonne-example}-c). When all sub-grids are generated by I-WFC (\ref{fig:Carcassonne-example}-d), each of them shares two edges with another adjacent sub-grids and are overlapped to form the final result (Fig. \ref{fig:Carcassonne-example}-f). Our algorithm can generate a grid of approximately 50,000 tiles within 10 seconds.

\begin{table}
\caption{One example of Carcassonne sub-complete tileset}
\label{table_example}
\centering
\begin{tabular}{c|c|c|c}
\hline
\multicolumn{4}{c}{tileset}\\
\hline
$(e_0, e_0, e_0, e_0)$ & $(e_2, e_0, e_2, e_0)$ & $(e_0, e_0, e_1, e_1)$ & $(e_2, e_2, e_0, e_0)$\\
$(e_0, e_1, e_0, e_1)$ & $(e_2, e_1, e_2, e_1)$ & $(e_0, e_0, e_2, e_2)$ & $(e_2, e_2, e_1, e_1)$\\
$(e_0, e_2, e_0, e_2)$ & $(e_2, e_2, e_2, e_2)$ & $(e_0, e_0, e_3, e_3)$ & $(e_2, e_2, e_3, e_3)$\\
$(e_0, e_3, e_0, e_3)$ & $(e_2, e_3, e_2, e_3)$ & $(e_1, e_1, e_0, e_0)$ & $(e_3, e_3, e_0, e_0)$\\
$(e_1, e_0, e_1, e_0)$ & $(e_3, e_0, e_3, e_0)$ & $(e_1, e_1, e_2, e_2)$ & $(e_3, e_3, e_1, e_1)$\\
$(e_1, e_1, e_1, e_1)$ & $(e_3, e_1, e_3, e_1)$ & $(e_1, e_1, e_3, e_3)$ & $(e_3, e_3, e_2, e_2)$\\
$(e_1, e_2, e_1, e_2)$ & $(e_3, e_2, e_3, e_2)$ & $(e_1, e_3, e_1, e_3)$ & $(e_3, e_3, e_3, e_3)$\\
\hline
\end{tabular}
\label{tab:carcassonne-sub-complete}
\end{table}

\subsubsection{Infinite Content}
\begin{figure}
\centering
\includegraphics[width=0.48\textwidth]{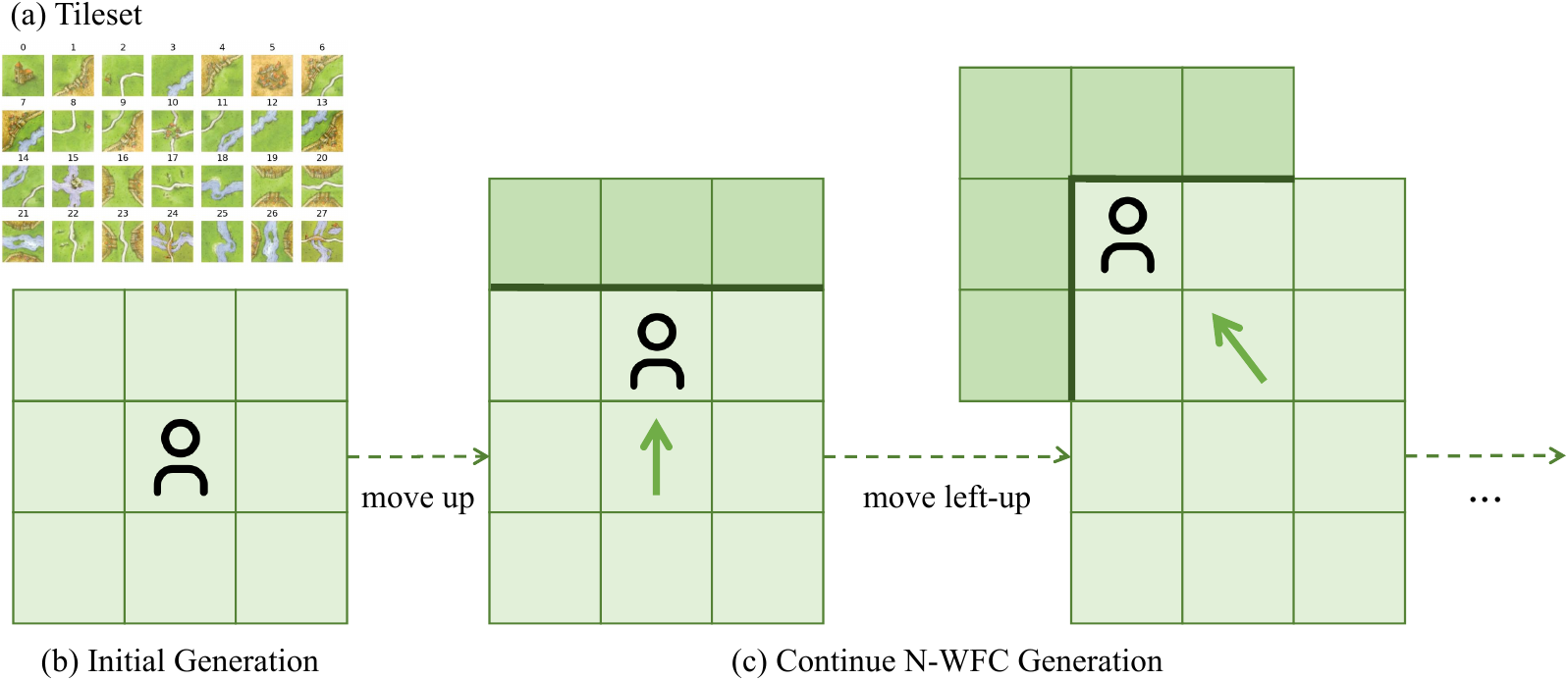}
\caption{Infinite game implementation with N-WFC and sub-complete tileset}
\label{fig:infinite-example}
\end{figure}

To generate infinite content, we modify the exterior generation process by taking the player as the origin and pre-generating the closest nine sub-grids of I-WFC with the player at the center (Fig. \ref{fig:infinite-example}-b). When the player moves to a different position, N-WFC detects any ungenerated sub-grids in the nine adjacent sub-grids whose origin is the new position and continuously generates new sub-grids using I-WFC (Fig. \ref{fig:infinite-example}-c,d).

Using the sub-complete tileset, I-WFC is guaranteed to return an accepted solution given two adjacent row and column constraints propagated from the pre-generated sub-grids. The deterministic characteristic of the generation process ensures that all the ranges that players have visited will not be backtracked. Since I-WFC is generated on a smaller grid, it completes the generation process in undetectable time, providing the player with an infinite feeling.

\subsection{Device Configuration}
The whole experiment was executed on a computer with  Microsoft Windows 11 Pro OS, Intel(R) Core(TM) 17-9750H CPU @ 2.60GHz, and 16GB RAM.

%% file: Content/6_Discussion.tex
\section{Discussion}

\subsection{Design Proprieties and Weight Brush System}
\begin{figure}
\centering
\includegraphics[width=0.45\textwidth]{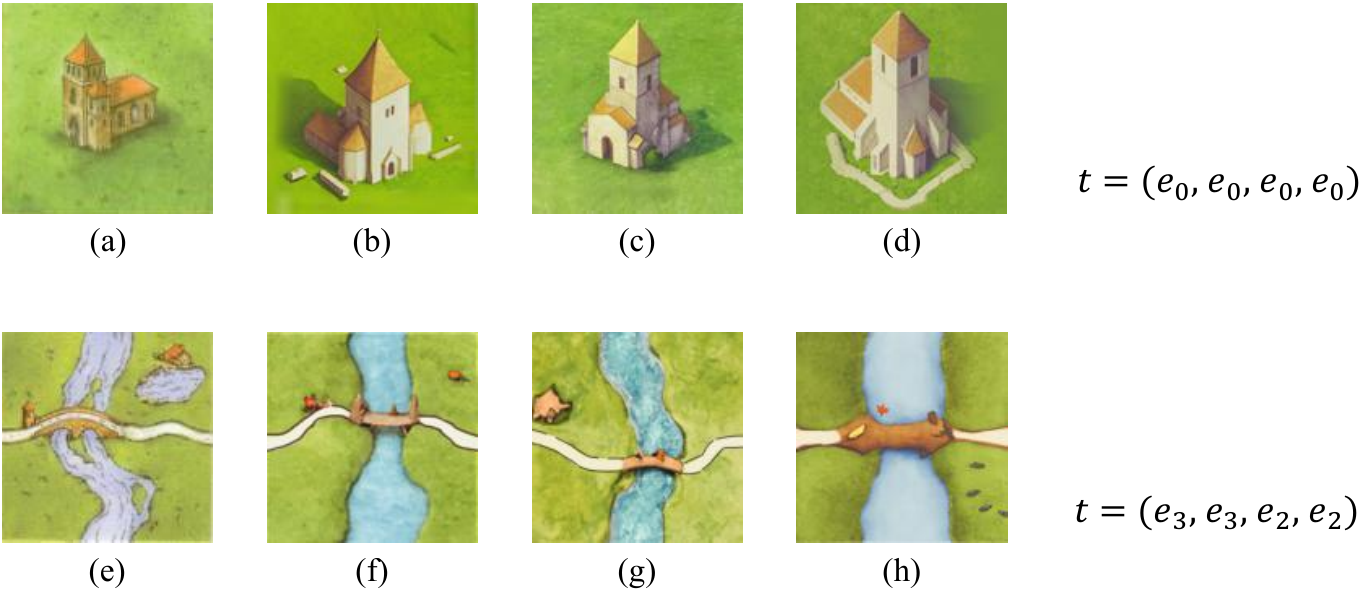}
\caption{Logically identical tiles have different design properties (Tile (a) and (e) are Carcassonne original tiles, while DALL·E generates the others)}
\label{fig:design-properties}
\end{figure}

In Section \ref{sec:complete-and-sub-complete}, we discussed the logical forms of sub-complete tilesets. However, tiles with identical logical forms can have different design properties, such as artistic expressions. We propose a \textit{Weight Brush System} that integrates design properties with the WFC algorithm, making the generation process more flexible. These design properties can be controlled during the generation process. As Fig. \ref{fig:design-properties} shows, (a)-(d) are identical tiles $t = (e_0, e_0, e_0, e_0)$ but with different artistic expression. In the design, we can think of tile (a) as an ordinary house and tile (d) as a luxury house. This is the same as tile (e) is more luxurious than tile (h). Tag \textit{luxurious} is a design property that can vary and share among different tiles.

The Weight Brush System consists of different weighted brushes, each representing tiles with a specific design property tag, such as buildings, landscapes, or quest points. Game designers can customize the weights of the brushes and use them to draw rough N-WFC-generated scenes. The weights affect the selection of a cell to collapse and the choice of which tile to collapse and can change the heuristic policy of the traditional WFC algorithm.

We argue that the weighted brush system is better in using I-WFC rather than a fine-grained diagonal generation process as it provides design-associated algorithms with selection strategies from being applied. It emphasizes game designers' priorities and produces maps with better design.

\subsection{Expand N-WFC and Sub-complete Tileset to 3D}
The approach of the N-WFC and sub-complete tileset can be easily extended to 3D. A 3D tile (voxel) consists of six edges (faces), denoted by $t = (e_n, e_s, e_w, e_e, e_u, e_d)$, which belong to three different face sets $\mathcal{E_{NS}}, \mathcal{E_{WE}}, \mathcal{E_{UD}}$. N-WFC operates on a grid $\mathbf{G} \subseteq \mathbb{Z}^3$ consisting of $M \times N \times O$ tiles. In the 3D case, N-WFC uses I-WFC with fixed $C \times C \times C$ sub-grids. The time complexity of N-WFC in 3D is $O(\frac{M\times N \times O}{C^3} d^{C^3} + (M \times N \times O) C^3 d^3)$. We can also add five new constraints to the definition of the sub-complete tileset.
\begin{flalign*}
    & \forall e_u, e_d \in \mathcal{E_{UD}}, \exists t \in \mathcal{T} \text{ s.t. } e_u(t) = e_u \wedge e_d(t) = e_d\\
    & \forall e_n \in \mathcal{E_{NS}}, \forall e_u \in \mathcal{E_{UD}},  \exists t \in \mathcal{T} \text{ s.t. } e_n(t) = e_n \wedge e_u(t) = e_u\\
    & \forall e_s \in \mathcal{E_{NS}}, \forall e_d \in \mathcal{E_{UD}},  \exists t \in \mathcal{T} \text{ s.t. } e_s(t) = e_s \wedge e_d(t) = e_d\\
    & \forall e_e \in \mathcal{E_{WE}}, \forall e_u \in \mathcal{E_{UD}},  \exists t \in \mathcal{T} \text{ s.t. } e_e(t) = e_e \wedge e_u(t) = e_u\\
    & \forall e_w \in \mathcal{E_{WE}}, \forall e_d \in \mathcal{E_{UD}},  \exists t \in \mathcal{T} \text{ s.t. } e_w(t) = e_w \wedge e_d(t) = e_d
\end{flalign*}
In the case of 3D, the sub-complete tileset also satisfies the three characteristics. The exterior of the N-WFC can use different generation methods to generate scenes that match the design expectations.

\subsection{Limitation and Future Work}
While we have demonstrated that the N-WFC of the simple tiled model using sub-complete tilesets can generate deterministic infinite content in a short time, the overlapping model has the advantage of automatically extracting the constraint relationship from an existing picture or model. Still, it currently fails to satisfy the sub-complete constraint. It creates large edge sets but is unable to extract enough tiles that meet the definition of the sub-complete tileset. This results in a lot of conflicts during N-WFC generation. Future work may consider modifying the overlapping model and implementing post-processing to meet the demands of sub-complete tilesets.

%% file: Content/7_Conclusion.tex
\section{Conclusion}
In this paper, we introduced Nested WFC (N-WFC), an optimization strategy for the WFC algorithm. N-WFC nests interior WFCs (I-WFC) into a large exterior generation process while maintaining the edge constraints, reducing exponential complexity to polynomial complexity. We proposed the concept of complete and sub-complete tilesets to maintain edge constraints during the N-WFC generation process and proved that under such tilesets, N-WFC can generate infinite, aperiodic, and deterministic content. We demonstrated through experiments that the combination of N-WFC and sub-complete tileset meets time complexity expectations. We provided an example of N-WFC's application in game development, specifically the Carcassonne game. We also proposed a weighted brush system to generate better scenes conforming to the design. Our work addresses the backtracking and conflicting shortcomings of WFC for large-scale and unlimited content generation, meets the needs of game designers, and is expandable and robust.